\documentclass[conference]{IEEEtran}

\IEEEoverridecommandlockouts
\usepackage{cite}
\usepackage{amsmath,amssymb,amsfonts}
\usepackage{algorithmic}
\usepackage{graphicx}
\usepackage{textcomp}
\usepackage{xcolor}
\usepackage{subcaption}
\usepackage[hidelinks,bookmarks=false]{hyperref}
\usepackage[letterpaper, top=0.75in, bottom=1.05in, left=0.625in, right=0.625in]{geometry}
\usepackage{array}
\usepackage{soul}

\def\BibTeX{{\rm B\kern-.05em{\sc i\kern-.025em b}\kern-.08em
    T\kern-.1667em\lower.7ex\hbox{E}\kern-.125emX}}
\newcommand{\im}[1]{\ensuremath{#1}}
\newcommand{\kw}[1]{\im{\mathtt{#1}}}
\newcommand{\THESYSTEM}{\kw{PART}}    
\begin{document}

\title{Multimodal Injury Risk and Performance Prediction in Tennis Using Weighted Ensemble Learning\\
{\large PART: Estimating Near-Term Athlete Readiness in Tennis}
\thanks{\textcopyright\ 2026 IEEE. Personal use of this material is permitted. Permission from IEEE must be obtained for all other uses, in any current or future media, including reprinting/republishing this material for advertising or promotional purposes, creating new collective works, for resale or redistribution to servers or lists, or reuse of any copyrighted component of this work in other works. Accepted author manuscript. Published in \emph{IEEE Systems, Man, \& Cybernetics Magazine} (Early Access), 2026. DOI: 10.1109/MSMC.2026.3685426.}}

\author{\IEEEauthorblockN{Weihao Qu, Dongyang Wang, Ling Zheng, Francisco E. Alvarez, Shobharani Polasa, and Jiacun Wang}
\IEEEauthorblockA{\textit{Department of Computer Science and Software Engineering} \\
\textit{Monmouth University}, West Long Branch, NJ 07764, USA \\
\{wqu, s1382037, lzheng, s1365567, s1365603, jwang\}@monmouth.edu}}

\newcommand{\wq}[1]{\textcolor{black}{#1}}

\maketitle

Machine learning has had a significant positive impact on the sports industry, with one of its most promising applications being the prediction of athlete performance and injury risk. Recent advances have employed state-of-the-art models to improve prediction accuracy, yet progress remains limited by data availability and the reliance on subjective observations or expert assessments. To address these limitations, researchers in sports such as soccer, basketball, and wrestling have begun integrating heterogeneous data sources, such as wearable device readings, with traditional subjective assessments. However, similar multimodal approaches remain underexplored in tennis.

In this work, we propose a multimodal weighted ensemble learning framework, \emph{Predictive Athlete Readiness for Tennis} ({\THESYSTEM}), to monitor athlete wellness and estimate near-term injury risk in tennis players. {\THESYSTEM} processes a wide range of inputs, including physiological metrics, training and match data, sleep information from wearable devices, self-reported questionnaires, vertical jump assessments, and motion analysis from match-play videos.
From these modalities, specialized machine learning and deep learning models independently extract four athlete-specific characteristics: overall wellness, injury risk, physical capability, and playing style.


To overcome the complexity of combining these diverse modalities, {\THESYSTEM} employs a supervised weighted ensemble integration strategy, assigning adaptive weights to each predictive model based on its reliability.
Evaluation on multimodal data collected from nine collegiate tennis players demonstrates that {\THESYSTEM} achieves strong performance in monitoring athlete wellness and estimating near-term injury susceptibility. Beyond collegiate athletes, the framework also shows promise for recreational tennis players, offering personalized insights to mitigate injury risk and optimize performance.

\section{Introduction}
{In professional sports, optimizing athletic performance while minimizing the risk of injury requires a comprehensive understanding of the many factors that impact an athlete's well-being, including physiological, psychological, training, and lifestyle factors\cite{adil2025metaverse}}.
A popular approach to studying how these factors affect athletes' physical health and injury risk involves applying machine learning to predict performance outcomes and potential injuries\cite{rossi2018effective}. Research in sports such as basketball, soccer, wrestling, and tennis has actively explored predictive modeling for performance and injury risk, identifying workout schedules, player characteristics, injury history, and workload as critical determinants~\cite{myers2020acute,Moreno_P_rez_2020, Liu_2023,Alvarez_2025, Guo_2024,deleeuw2022modeling}.

 Among these sports, tennis presents unique challenges and opportunities:
 (1) Tennis holds a prestigious place globally, with widespread popularity and a diverse fan base.
(2) Tennis is complex, requiring a blend of athleticism, strategic thinking, and mental resilience, which makes performance and injury prediction especially challenging.
(3) Overuse and repetitive motion injuries are common among both professionals and recreational players~\cite{Abrams_2012,Kaiser_2021}, highlighting the need for robust prediction tools that can mitigate risk and enhance performance.

Current machine-learning-based approaches in tennis primarily rely on official match statistics, subjective observations, and expert assessments. {This leads to a common challenge in existing research in sports performance and injury prediction: many models rely heavily on subjective observations and expert assessments~\cite{Huang_2022,fan2026fixedtime, zhang2025sampleddata}, limiting their ability to capture the complex interplay of various factors. To overcome these limitations, recent work has turned to richer data sources.} Wearable devices can objectively track physiological signals, workload, and recovery status~\cite{Qu_2016, Sathyanarayana_2016,  Yu_2024}. Athlete self-report measures (ASRM) provide valuable subjective insights into mental and physical wellness~\cite{saw2017athlete}. Video analysis, leveraging advances in computer vision, can reveal biomechanical patterns and technical playing styles across training and competition~\cite{Huang_2025}. These multimodal data streams collectively offer a more holistic picture of athlete health and readiness.

\wq{However, the integration of multimodal data also introduces significant complexity. {Each data source captures distinct but complementary aspects of player status, and naïvely combining them can dilute signal quality or overweight less reliable modalities. This challenge motivates the use of weighted ensemble learning, which provides a principled framework for fusing heterogeneous models~\cite{kordestani2025condition, xiao2024fixedtime}.By assigning modality-specific weights that reflect predictive reliability and contextual importance, weighted ensembles can balance contributions from diverse data sources, reduce noise, and capture the nuanced interplay of factors influencing injury and performance~\cite{sun2025blockchain}}}
 

\wq{To this end, we propose a multimodal weighted ensemble learning framework,  \textbf{P}redictive \textbf{A}thlete \textbf{R}eadiness framework for \textbf{T}ennis ({\THESYSTEM}) for assessing both performance and injury risk in tennis players. {\THESYSTEM} processes multimodal data from collegiate athletes, including wearable-derived physiological and workload measures, self-reported survey responses, and automated video-based motion analysis.}
Our framework employs a two-stage design: in the Separating stage, specialized machine learning models (XGBoost regressors/classifiers, MLP, LSTM, and computer vision models) independently predict modality-specific outcomes such as wellness, repetition fatigue, physical capability, and technical playing style. In the Integration stage, these outputs are combined using supervised weighted ensemble learning to generate a comprehensive Athlete Readiness Score (ARS). 
Case studies demonstrate that {\THESYSTEM} effectively monitors athlete condition and provides interpretable insights, identifying athletes at elevated injury risk and guiding actionable interventions such as load management and targeted recovery strategies. By leveraging multimodal data and weighted ensemble learning, our framework offers a robust tool for both professional and recreational tennis players seeking to optimize performance while reducing injury risk.

\begin{figure*}[!h]
\centering
\includegraphics[width=0.9\textwidth]{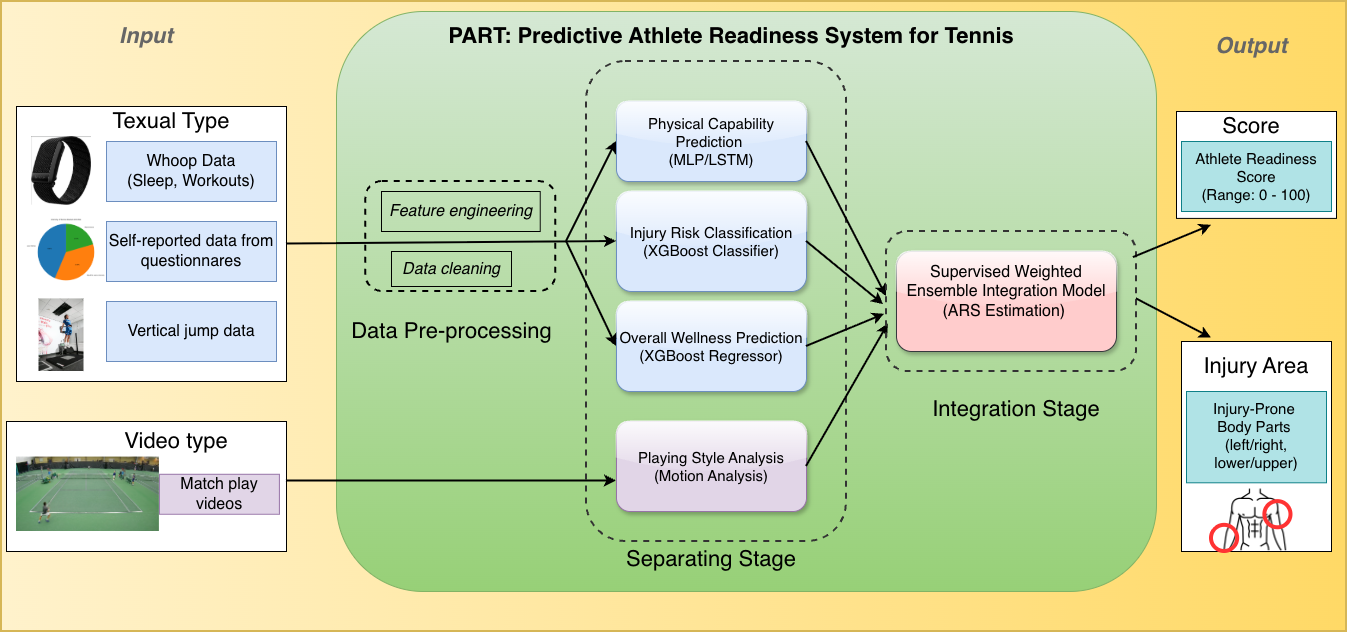}
\caption{System architecture of {\THESYSTEM}.}
\label{fig:system_architecture}
\end{figure*}

\section{Methodology}
\label{sec:arch}
\subsection{System Architecture}

The overall architecture of {\THESYSTEM} is illustrated in Figure~\ref{fig:system_architecture}, consisting of (1) the data input layer, (2) pre-processing modules, (3) the separating stage with four models learning the physical capability, injury risk, overall wellness and playing style of the tennis athletes, and (4) the integration stage with an integration model to generate the outputs.
This modular design allows for flexibility and scalability in processing diverse data types and integrating new models.

The data collection of the input of {\THESYSTEM} stems from four primary sources: WHOOP wearable devices providing sleep and workout data, self-reported data from questionnaires, and vertical jump data from profession physicians and videos of match play. The first three are textual-type data with a rigorous preprocessing phase, including data cleaning and feature engineering, we elaborate the details in Section~\ref{sec:data}.

 The \textbf{Seperating} stage 
consists of four specialized predictive models: (1) A Physical Capability Prediction Model utilizes MLP and LSTM architectures to forecast an athlete's physical readiness based on temporal sequences of data; (2) An Injury Risk Classification Model employs an XGBoost Classifier to assess the probability of injury, leveraging both physiological and self-reported data; (3) An Overall Wellness Prediction Model relies on an XGBoost Regressor to evaluate the athlete's general well-being by integrating multiple data features.
(4) A playing style model to predict the athlete's style based on match play videos. We unveil more details in Section~\ref{sec:model}.

In the  \textbf{Integeration} stage, our supervised weighted ensemble integration model fully utilizes previous results of these four models in separating stage, to calculate an \textit{Athlete Readiness Score (ARS)}. 
ARS defined in {\THESYSTEM} is a composite score ranging from 0 to 100, where higher scores indicate better overall readiness for performance. It provides a holistic assessment of an athlete’s condition, balancing physical capability, wellness, and injury risk.

\section{Data Collection and Preprocessing}
\label{sec:data}

\subsection{Participants}
Data were collected from nine collegiate tennis players (Male: 5, Female: 4; Age: 20.3 ± 1.5 years; Height: 175.2 ± 8.7 cm; Weight: 68.9 ± 9.2 kg) over a period of 16 weeks. All participants were informed about the study procedures and provided written consent.
\subsection{Data Collection}
\begin{figure}[h]
\centering
\begin{subfigure}{0.45\textwidth}
\includegraphics[width=\linewidth]{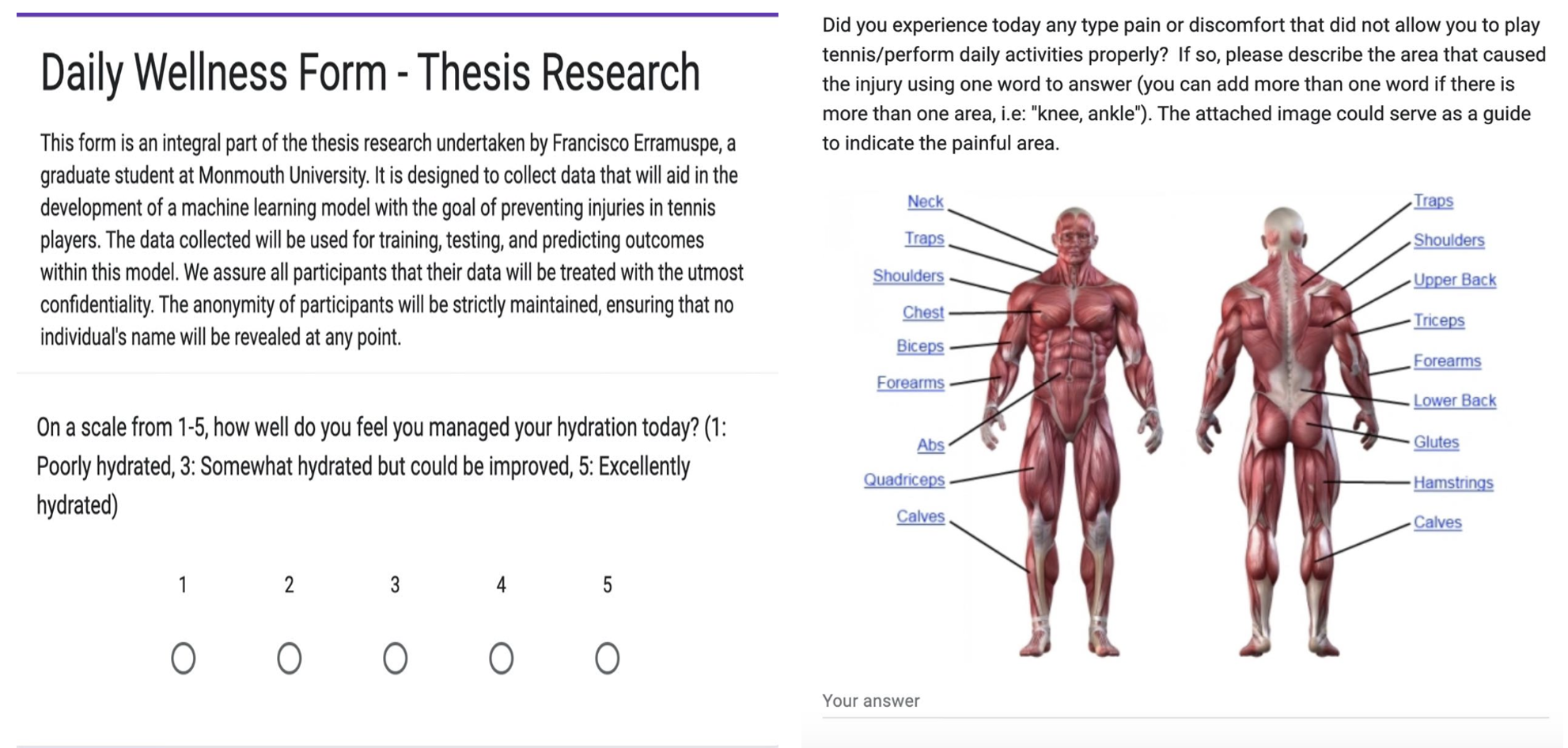}
\caption{Questionnaire}
\end{subfigure}
\hfill
\begin{subfigure}{0.18\textwidth}
\includegraphics[width=\linewidth]{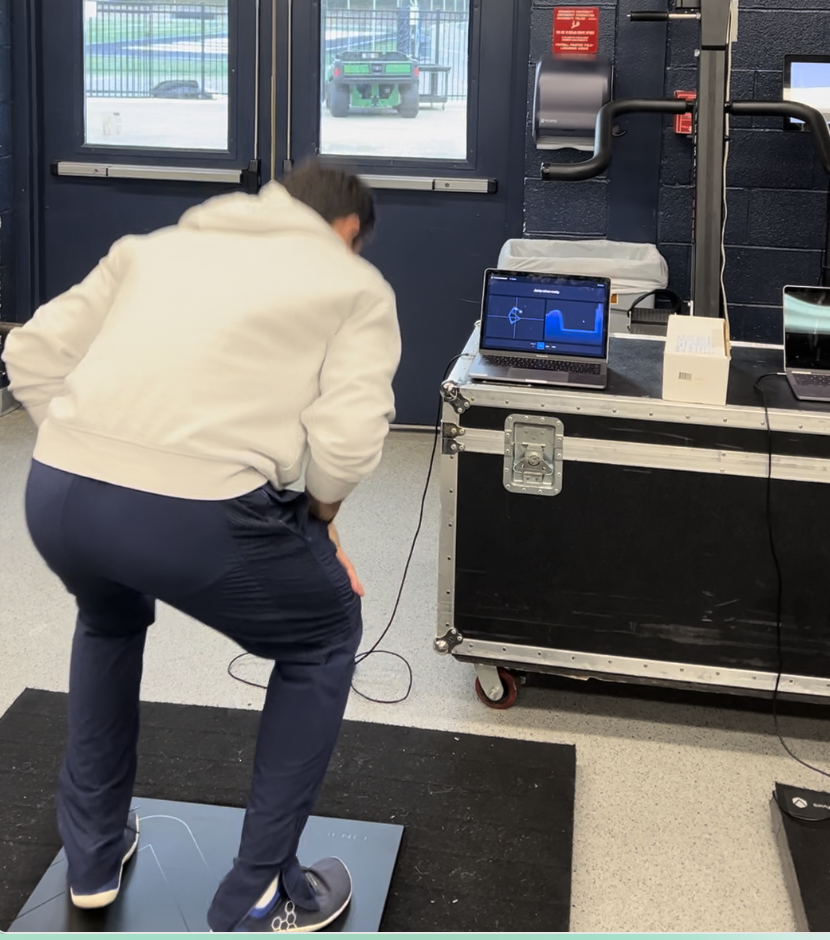}
\caption{Sparta jump}
\end{subfigure}
\hfill
\begin{subfigure}{0.32\textwidth}
\includegraphics[width=\linewidth]{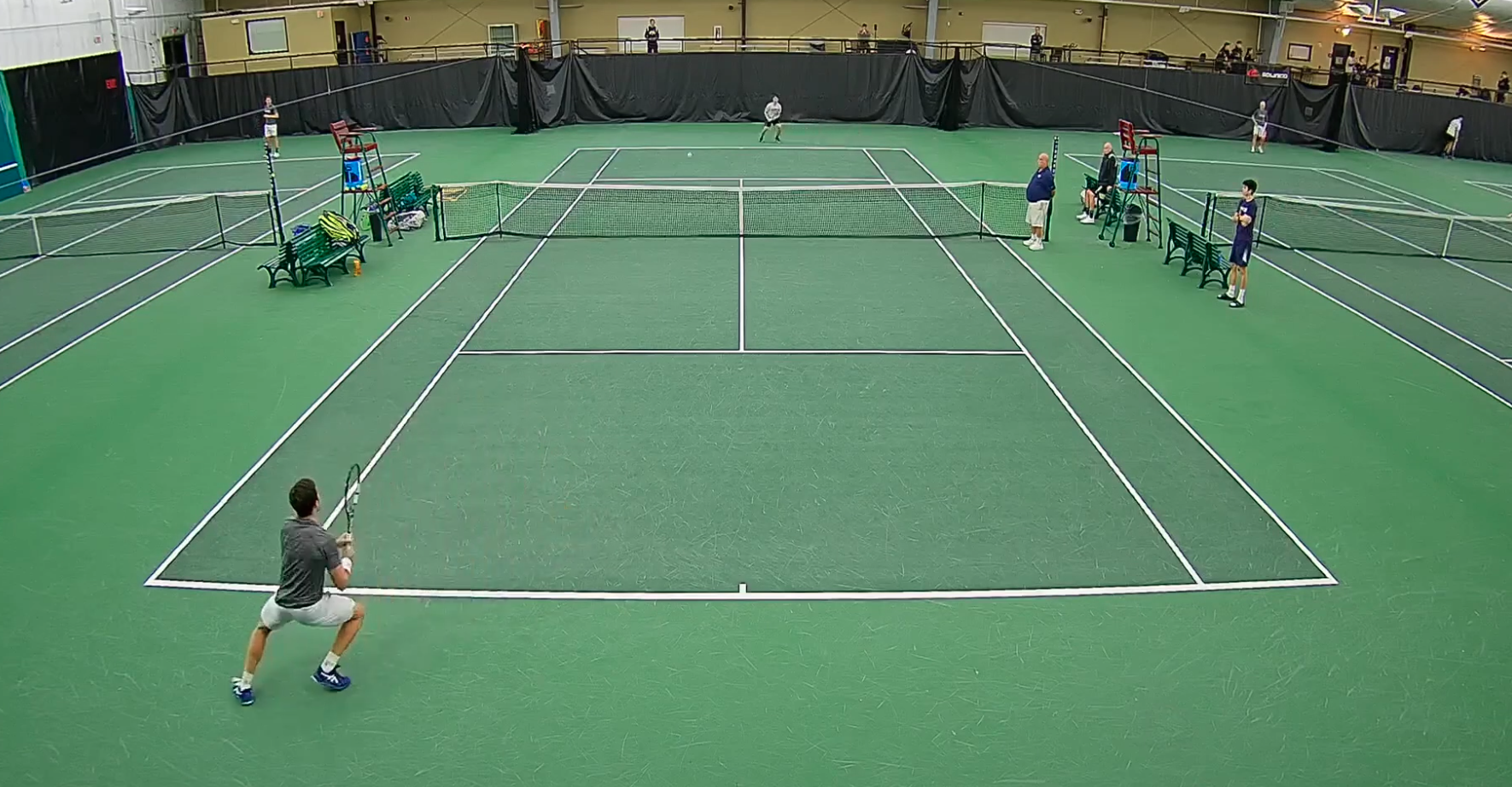}
\caption{Match play video}
\end{subfigure}
\caption{Examples of author-controlled inputs: (a) questionnaire; (b) vertical jump test; and (c) match-play video. WHOOP participant data are shown in Figure~\ref{fig:sampledata}(b).}
\label{fig:questionnaire}
\end{figure}

\begin{figure}[h]
\centering
\begin{subfigure}{0.45\textwidth}
\includegraphics[width=\linewidth]{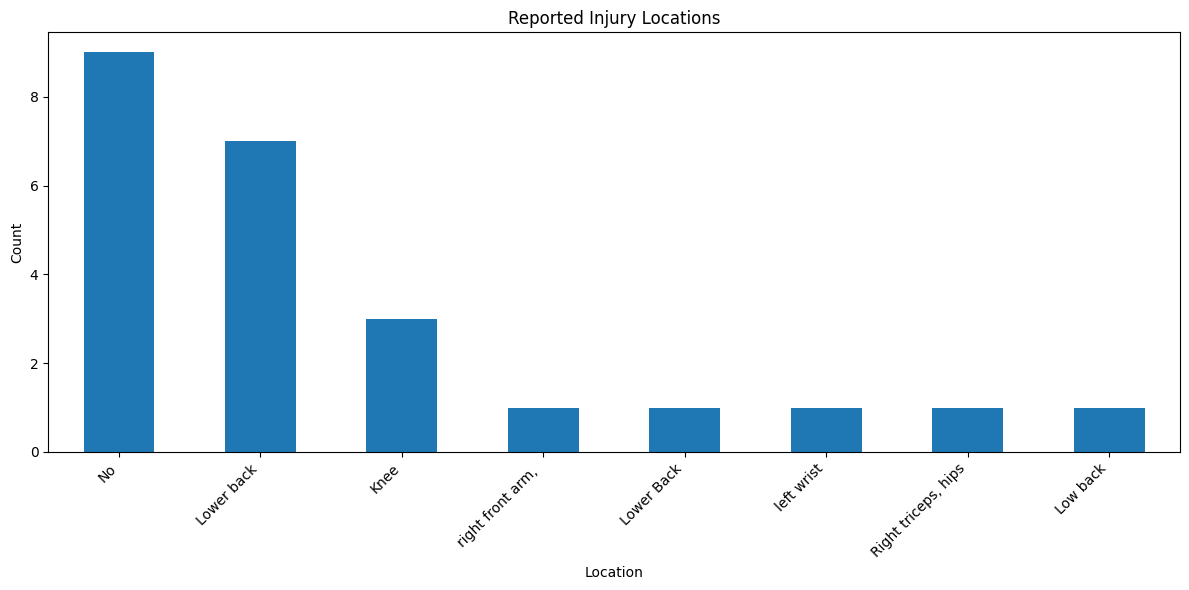}
\caption{Questionnaire Result}
\end{subfigure}
\hfill
\begin{subfigure}{0.45\textwidth}
\includegraphics[width=\linewidth]{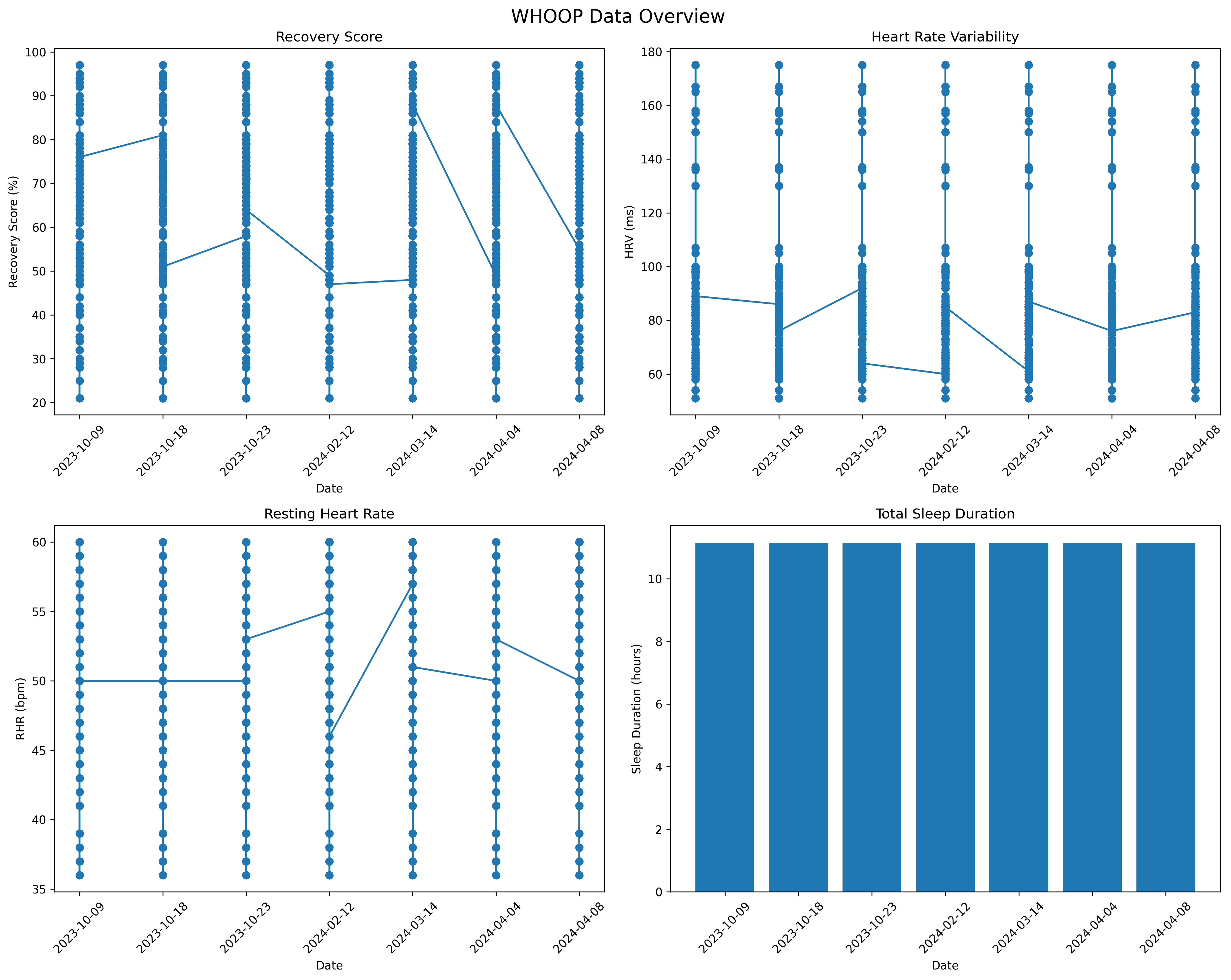}
\caption{WHOOP sample data}
\end{subfigure}
\hfill
\begin{subfigure}{0.45\textwidth}
\includegraphics[width=\linewidth]{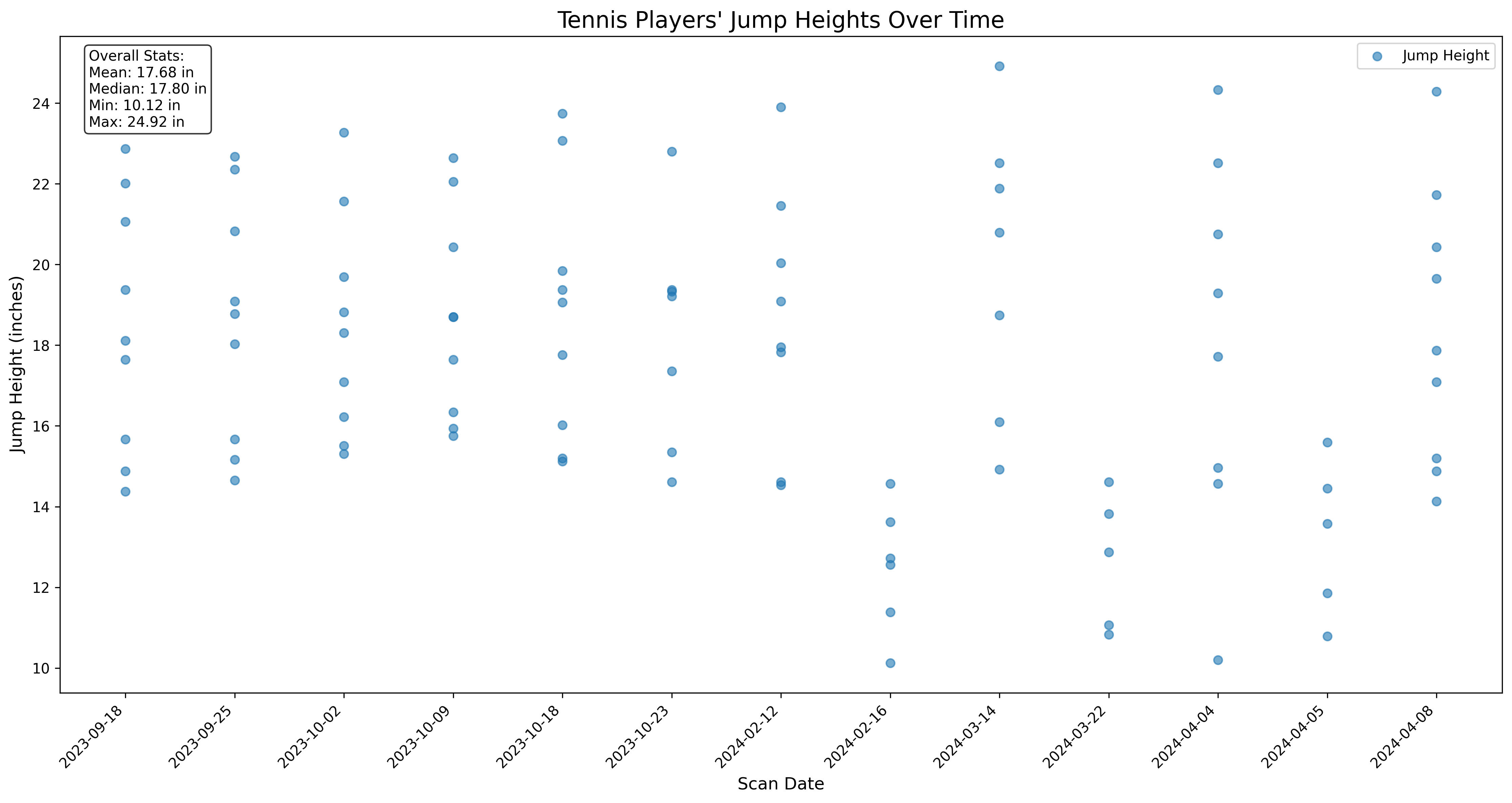}
\caption{Sample Jump data}
\end{subfigure}
\caption{Collected Data Sample: (a) The result of questionnaire on reported injury locations from participants (b) Sample of the sleep, heart rate, recovery score tracked by WHOOP device (c) The Vertical jump data sample from Sparta testing.}
\label{fig:sampledata}
\end{figure}

Data were gathered from four distinct sources. Figure~\ref{fig:questionnaire}(a) shows the daily online \emph{questionnaire} tracking self reports of all participants. We collected responses to 10 questions addressing training load, physical discomfort and pain, mental stress, nutrition, and hydration. Additionally, the questionnaire allowed participants to specify areas of discomfort. Figure~\ref{fig:sampledata}(a) records the reported injury locations of the participants during their training and matches, which is the main source of injury-area prediction in our framework.

We use WHOOP devices to record objective sleep data, continuous physiological data, and detailed workout data encompassing training sessions and matches for our participants. WHOOP devices are widely used in professional tennis~\cite{gupta2025whoop}. Sample data on participants' sleep, recovery, and heart rate tracked by WHOOP devices are shown in Figure~\ref{fig:sampledata}(b).

We collected vertical jump data from our participants through standardized weekly or bi-weekly Sparta testing organized at Monmouth University. In Figure~\ref{fig:questionnaire}(b), one collegiate tennis player from the Monmouth University tennis team is recording the date and jump height in inches. We present sample jump data in Figure~\ref{fig:sampledata}(c).

The match play videos of our participants were collected by the Monmouth University tennis team during the spring 2024 season; a snapshot of one match is shown in Figure~\ref{fig:questionnaire}(c).

\subsection{Feature Engineering}

We cleaned the data from questionnaires, WHOOP devices, and Sparta testing by standardizing missing values, converting data types, imputing missing values. The cleaned textual data contains $85$ attributes and $82804$ entries.  
Our feature engineering select the following key physiological metrics among the $85$ attributes:

\begin{itemize}
    \item \textbf{Heart Rate Variability (HRV)}: Measures the variation in time between heartbeats, showing stress and recovery.
    \item \textbf{Resting Heart Rate (RHR)}: A lower resting heart rate suggests better fitness and recovery.
    \item \textbf{Sleep Efficiency}: The percentage of time spent asleep while in bed, indicating recovery quality.
    \item \textbf{Activity Strain}: A measure of the intensity of physical activity performed by the athlete.
\end{itemize}

To address skewed distributions and satisfy linear regression assumptions, we applied Logarithmic Transformations to variables such as ``Heart Rate Variability (ms)'' to help normalize the data distribution.
To enable the model to capture non-linear relationships between predictors and the target variable, we generated Polynomial Features. This technique expands the feature space, allowing for more complex relationships to be modeled. 

\section{Separating Stage: Model Development}
\label{sec:model}
In the Separating stage, we developed different models for textual-type data and video-type data to learn the wellness, injury risk, injury area, physical capability and playing style of the tennis athletes.
 The performance of these models was evaluated using appropriate metrics for each task. 

\subsection{Overall Wellness Prediction (Regression Models)}
The goal of Overall Wellness Prediction is to estimate the Recovery Score, which ranges from 0 to 100 and reflects an athlete’s recovery from physical exertion. 
For this purpose,
 regression models including Linear Regression (with various enhancements such as feature transformation and interaction terms), Polynomial Regression, Lasso Regression, and XGBoost Regressor are trained using input variables such as heart rate, strain, and sleep metrics. Models are validated using 3-fold group cross-validation (6:3 player split) with 95\% confidence intervals from 100 bootstrap iterations to ensure robust performance estimates.

\begin{table}[h]
\caption{Regression Performance (Overall Wellness)}
\label{tab:regression_results}
\centering
\begin{tabular}{lcc}
\hline
\textbf{Model} & \textbf{MAE} & \textbf{R\textsuperscript{2} Score} \\
\hline
Baseline Linear Regression           & $16.81 \pm 12.48$ & $-0.81 \pm 2.50$ \\
Linear Regression  & $21.62 \pm 19.54$ & $-2.89 \pm 3.62$ \\
Polynomial Regression                 & $316.91 \pm 454.91$ & $-1688.51 \pm 2009.61$ \\
Lasso Regression                      & $10.72 \pm 2.78$ & $0.38 \pm 0.23$ \\
Gradient Boosting                     & $9.76 \pm 1.00$ & $0.43 \pm 0.10$ \\
XGBoost                      & $8.04 \pm 1.87$ & $0.65 \pm 0.12$ \\
\hline
\multicolumn{3}{l}{\footnotesize 95\% CI shown as mean $\pm$ margin of error.}
\end{tabular}
\end{table}

\begin{figure}[h]
\centering
\begin{subfigure}{0.23\textwidth}
\includegraphics[width=\linewidth]{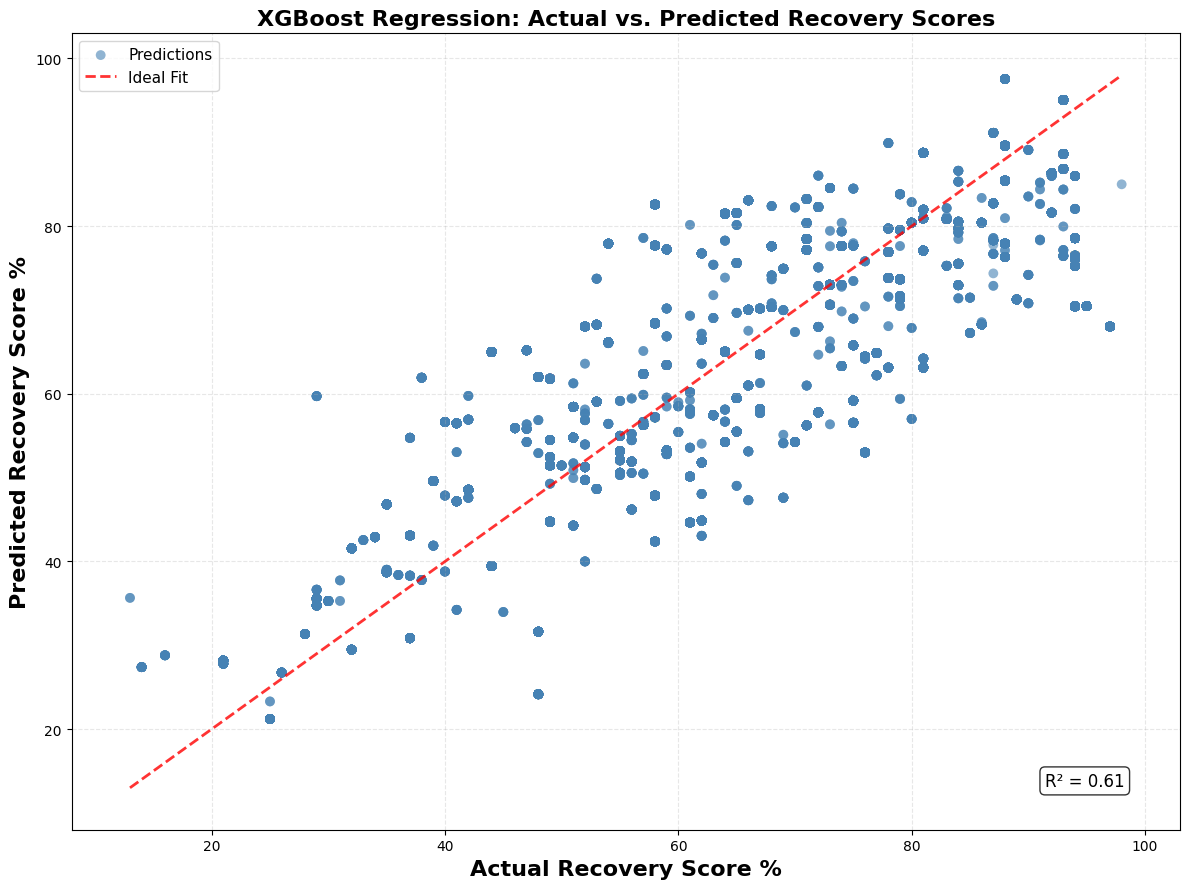}
\caption{{XGBoost Regression: Actual vs. Predicted Recovery Scores}}
\end{subfigure}
\hfill
\begin{subfigure}{0.24\textwidth}
\includegraphics[width=\linewidth]{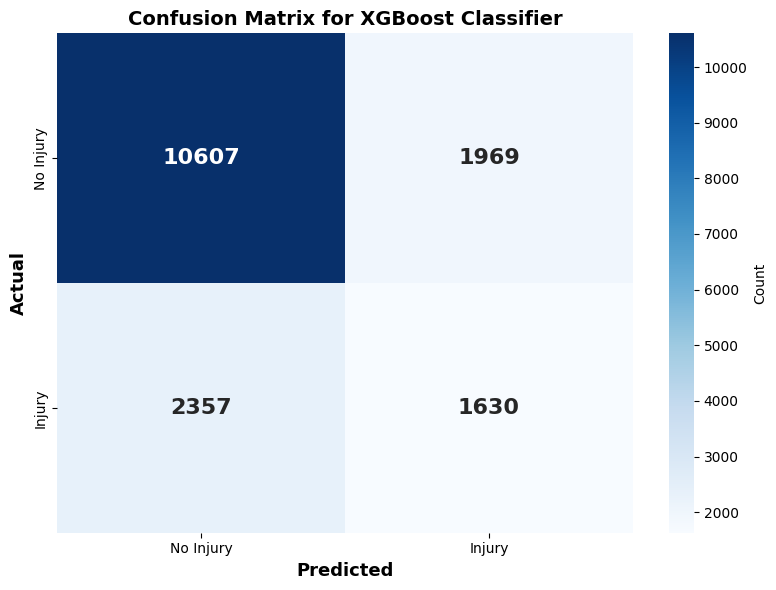}
\caption{{Confusion Matrix for XGBoost Classifier}}
\end{subfigure}
\caption{Regression and classifier results. (a) XGBoost regression: actual versus predicted recovery scores. The $R^2$ value shown reflects a representative test fold; the cross-validated mean is $0.65 \pm 0.12$ (Table~\ref{tab:regression_results}). (b) The confusion matrix for the XGBoost classifier.}
\label{fig:rcresults}
\end{figure}

The performance metrics of these models are summarized in Table~\ref{tab:regression_results}, which reports the Mean Absolute Error (MAE), and the coefficient of determination (R\textsuperscript{2}  Score). 
The XGBoost Regressor outperformed all linear regression models, achieving an R\textsuperscript{2} score of 0.65. Based on the results, the three most important features are player-normalized Heart Rate Variability (HRV) (0.496), sleep performance (0.160), and raw HRV (0.132). Figure \ref{fig:rcresults}(a) illustrates the relationship between actual and predicted recovery scores using the XGBoost Regression model. This demonstrated that XGBoost is highly effective at capturing complex non-linear interactions within physiological and performance data. Our framework uses XGBoost regression model for the overall wellness of target athletes. 




\subsection{Injury Risk Classification (Classification Models)}

The goal of Injury Risk Classification is to predict whether an athlete is at High Risk or Low Risk of injury within a specific time frame. Injury risk is defined as a binary label (1/0) based on the daily questionnaire.The classification was based on physiological metrics such as Activity Strain, HRV, and Resting Heart Rate, as well as self-reported survey responses.

We implemented Logistic Regression, XGBoost Classifier, Decision Tree Classifier, and Random Forest Classifier. These models were tasked with predicting the likelihood of an athlete sustaining an injury based on the collected data. Table \ref{tab:classification_results} summarizes their performance.

\begin{table}[h]
\caption{Classification Performance (Injury Risk)}
\label{tab:classification_results}
\centering
\begin{tabular}{lccc}
\hline
\textbf{Model} & \textbf{AUC-ROC} & \textbf{Accuracy} & \textbf{F1 Score (Injury)} \\
\hline
Logistic Regression      & 0.61 $\pm$ 0.01 & 0.48 $\pm$ 0.09 & 0.40 $\pm$ 0.01 \\
XGBoost         & 0.64 $\pm$ 0.01 & 0.74 $\pm$ 0.01 & 0.43 $\pm$ 0.01 \\
Decision Tree & 0.65 $\pm$ 0.01 & 0.74 $\pm$ 0.01 & 0.43 $\pm$ 0.02 \\
Random Forest  & 0.64 $\pm$ 0.01 & 0.74 $\pm$ 0.01 & 0.43 $\pm$ 0.01 \\
\hline
\multicolumn{4}{l}{\footnotesize 95\% confidence intervals shown as mean $\pm$ margin of error.}
\end{tabular}
\end{table}

The XGBoost, Decision Tree, and Random Forest classifiers have similar results in terms of area under the curve (AUC)-receiver operating characteristic (ROC) and F1 score, showing their abilities to distinguish between athletes at risk of injury and those who are not. The comparative high accuracy across these three models, coupled with low F1 Scores for injury class, indicates challenges in predicting the minority class due to dataset imbalance,  with injury cases representing approximately 24\% of the total observations. To address class imbalance, we employed techniques such as adjusting class weights in the XGBoost Classifier and ensuring stratified sampling for training and testing.
Figure~\ref{fig:rcresults}(b) presents the confusion matrix for the XGBoost Classifier, highlighting its performance in identifying non-injury cases and its limitations in correctly predicting injury occurrences. {The three most important features for injury risk prediction are HRV (0.334), Resting Heart Rate(RHR) (0.184), and Recovery Score (0.112).}





In addition to overall injury risk, we developed separate XGBoost models for upper-body and lower-body injury predictions. 
 These models demonstrated improved specificity, allowing for targeted preventive measures based on the athlete's specific injury risk profile. \wq{A comparison of upper-body and lower-body injuries reveals that the model achieves an AUC-ROC of 0.712 for upper-body injuries and 0.703 for lower-body injuries.}

\subsection{Physical Capability Prediction}
\label{PCP}

The athlete’s physical capability was quantified using the Performance Capability Score (PCS), which was derived as a weighted combination of multiple readiness indicators. The PCS integrates Reactive Strength (RS) and Vertical Jump Performance Score (VJPscore) from jump data, Sleep Efficiency (SE), Heart Rate Variability (HRVscore), and Sleep Score (SS) from the WHOOP device, as well as a Quality Score (QS) obtained from athlete questionnaires.

{We developed models using both Multi-Layer Perceptron (MLP) and Long Short-Term Memory (LSTM) architectures to predict the athlete's physical capability score based on a 7-day sequence of physiological and activity metrics. We implemented several overfitting mitigation strategies: (1) Dropout after each LSTM layer, (2) Early stopping with patience, (3) 7-day sequence length balancing temporal context with model complexity, and (4) 3-fold Group cross-validation with player-based splits.

The LSTM substantially outperformed the MLP, with an RMSE of 0.749 compared to 2.506 for the MLP, and an R² of 0.993 versus 0.922, demonstrating the LSTM's superior ability to exploit temporal dependencies in the 7-day input window. The strong test set performance (R² = 0.993, MAE = 0.248) confirms that the implemented regularization strategies effectively prevented overfitting. On the held-out test set, predicted physical capability scores spanned approximately 60–78\%, providing a nuanced view of athlete readiness across different days and conditions.}


\subsection{Playing Style Analysis}
\label{PSA}
To enhance our understanding of individual playing patterns and their impact on athlete wellness, we implemented an automated video analysis system to track and analyze match play. 
\wq{One way to quantify rally dynamics in tennis is via Average Shots per Point (ASP), i.e. the mean number of strokes (including the serve and return) exchanged until the point ends. In tennis performance analysis area, ASP can serve as a useful metric for comparing match style, player consistency, surface effects, and strategy~\cite{Prieto_Lage_2023}.} 
 Based on the ASP metric, players are classified into four distinct categories:

\begin{itemize}
    \item \textbf{Highly Aggressive} (ASP $\leq$ 5): Players who consistently end points quickly through powerful shots
    \item \textbf{Aggressive} (5 $<$ ASP $\leq$ 8): Players who maintain an attacking style while showing more rally tolerance
    \item \textbf{Neutral} (8 $<$ ASP $\leq$ 12): Players who balance aggressive and defensive play
    \item \textbf{Defensive} (ASP $>$ 12): Players who primarily rely on consistency and extended rallies
\end{itemize}

\subsection{Discussion}
\paragraph{Model Choice} In the separating stage, different models are selected based on the specific factors being analyzed. XGBoost Regressor is employed for its ability to handle non-linear relationships in physiological data and its computational efficiency, making it suitable for real-time predictions. XGBoost Classifier is chosen for injury risk classification due to its strong performance on imbalanced datasets. To analyze physical capability, LSTM and MLP models are utilized for their ability to process temporal sequences of training data effectively. 


 \paragraph{Correlation Between Physical Capability and Injury Probability} Lower Heart Rate Variability (HRV) indicated higher stress levels and inadequate recovery. Elevated Resting Heart Rate (RHR) signaled fatigue or overtraining. 
 Higher activity strain correlated with greater physical stress.

These correlations underscore the importance of monitoring physiological indicators to proactively identify athletes at risk of injury. By integrating these metrics into our predictive models, coaches and trainers can implement targeted interventions to mitigate injury risks.

\section{Integration Stage: Model Integration and Output Generation}
\label{sec:integration}

\subsection{ARS Labelling}
\label{sec:ars_labelling}

To enable supervised learning to determine the optimal weights in ARS, we first established meaningful labels for the ARS based on expert evaluations. Coaches and sports scientists assessed each athlete's overall readiness using a comprehensive scale from 0 to 100. This assessment considered various factors, including physical performance metrics, recovery status, training load, and subjective wellness reports.

\subsection{Weight Determination via Supervised Learning}
\label{sec:weight_determination}


We wanted the model to learn the optimal weights 
\( w_1, w_2, w_3 \) in the Athlete Readiness Score (ARS), corresponding to the Overall Wellness Score (OWS), Injury Risk (IR), and Physical Capability Score (PCS), respectively. {A linear weighting scheme was adopted for its simplicity, transparency, and interpretability, while still capturing modality importance.}
 



These weights were determined through supervised learning using historical data from September to December 2023. This approach ensures the weights reflect actual relationships between metrics and athlete performance.
Weekly data collection involved: WHOOP physiological metrics, Vertical jump test results, Daily wellness questionnaires, Coach evaluations, and Match performance data. 

The training process yielded final weight values of $w_1 = 0.35$, $w_2 = 0.30$, and $w_3 = 0.35$, which were used in the initial ARS calculation.

\subsection{Incorporating Video Analysis Results}
\label{sec:video_incorporation}


The playing style from the video-type data serves as an important weighting factor in our ARS calculations, with higher weights assigned to recovery metrics for defensive players who typically experience greater physical demands during matches. 

The ASP metric is integrated into our framework according to Section IV-D. The player’s style classification derived from video analysis determines the corresponding weighting factor $w_{\text{style}}$. Specifically, $w_{\text{style}}$ is set to 1.0 for highly aggressive players, 1.2 for aggressive players, 1.4 for neutral players, and 1.6 for defensive players. The final ARS is then adjusted by dividing the initial ARS by the $w_{\text{style}}$.



This adjustment ensures that our wellness predictions account for the varying physical demands associated with different playing styles, providing more accurate and personalized assessments of athlete readiness and injury risk.

\subsection{ARS Computation}

\begin{table}
\centering
\caption{Sample Training Dataset (One Week Period)} 
\label{tab:training_data}
\begin{tabular}{|c|c|c|c|c|c|c|}
\hline
\textbf{Player}  & \textbf{Overall} & \textbf{Injury} & \textbf{Physical} & \textbf{Style} & \textbf{Final} \\
\textbf{ID} & \textbf{Well (OWS)} & \textbf{Risk (IR)} & \textbf{Cap (PCS)} & \textbf{Wt} & \textbf{ARS} \\ \hline
P1 &  82 & 0.25 & 75 & 1.2 & 64.54 \\ \hline
P2 &  78 & 0.30 & 72 & 1.4 & 52.50 \\ \hline
P3 &  85 & 0.15 & 82 & 1.0 & 83.95 \\ \hline
P4 &  75 & 0.30 & 65 & 1.6 & 43.75 \\ \hline
P5 &  88 & 0.15 & 80 & 1.2 & 70.25 \\ \hline
P6 &  70 & 0.35 & 60 & 1.4 & 46.43 \\ \hline
P7 &  95 & 0.05 & 90 & 1.0 & 93.25 \\ \hline
P8 &  65 & 0.40 & 55 & 1.6 & 37.50 \\ \hline
P9 &  82 & 0.22 & 72 & 1.2 & 64.42 \\ \hline
\end{tabular}
\end{table}
Table \ref{tab:training_data} shows the training dataset used for supervised learning. The supervised learning implementation used scikit-learn's LinearRegression model. 



\wq{To interpret the ARS scores within our framework, we apply the following thresholds: players with scores above 80 are considered to be in good condition with a low risk of injury; scores between 60 and 80 indicate a standard physical condition; and scores below 60 suggest suboptimal condition and a high risk of injury. In such cases, players are advised to undergo further evaluation, reduce training loads, or adjust match schedules.}

As shown in Table \ref{tab:training_data}, player P7 stood out with a final ARS score of 93.25, reflecting excellent wellness, minimal injury risk, and strong physical capacity. In contrast, four players (P2, P4, P6 and P8) scored below 60, a level that generally indicates accumulated fatigue and the need for additional rest or recovery before resuming full training loads. In particular, P8, with the lowest score of 37.50, fell into a dangerous range, suggesting the highest injury risk compared to the other athletes and requiring immediate measures such as medical evaluation and enforced recovery.

\section{ Evaluation via Case Studies}
\label{casestudy}
To illustrate our ARS system's monitoring capabilities in real scenarios, we present two case studies that demonstrate its ability to capture both acute injuries and cumulative fatigue. It is important to note that the current results reflect monitoring and near-term risk estimation rather than validated prospective injury forecasting. Given the practical challenges of longitudinal athlete tracking and the pilot nature of this study with a limited cohort of nine players, these illustrative case studies provide qualitative evidence of the framework's monitoring capabilities and demonstrate feasibility for future large-scale validation.

\subsection{Case Study 1: Non-Fatigue Injury Analysis}
\begin{figure}[h]
\centering
\includegraphics[width=0.48\textwidth]{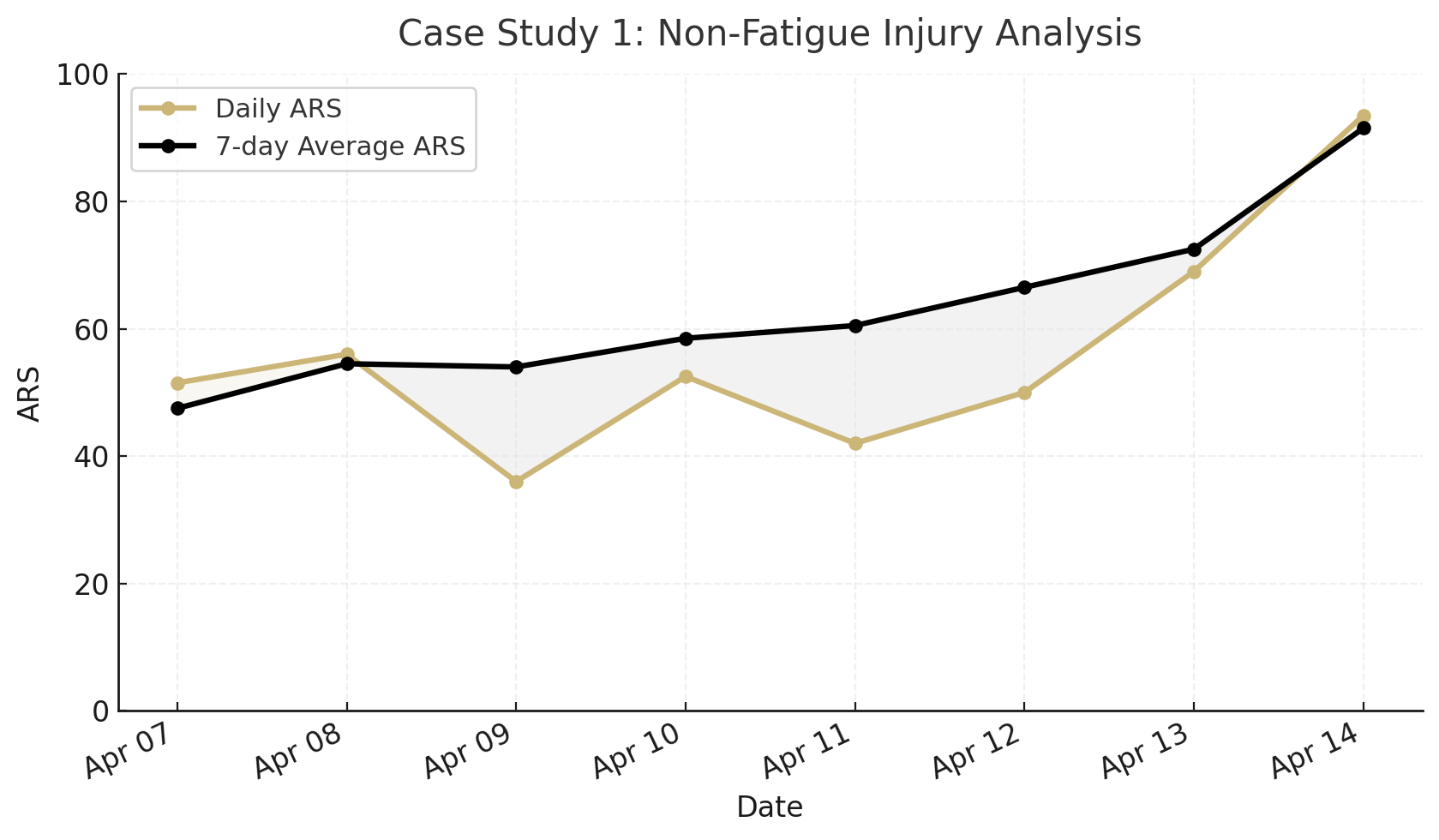}
\caption{ARS tracking during and after ankle sprain (April 2024). The plot shows a increasing trend in ARS score during recovery period.}
\label{fig:case_study_1}
\end{figure}

Figure~\ref{fig:case_study_1} presents an interesting case that demonstrates both limitation and strength of our framework. The data captures an ankle sprain injury event and its aftermath in April 2024.

This case highlights an important distinction in sports injuries: while some injuries are driven by fatigue and can be anticipated through physiological and performance metrics, others, such as ankle sprains, often result from acute mechanical incidents rather than accumulated stress. The ankle sprain sustained on April 11, 2024, falls into the latter category, representing an injury type that is inherently difficult to predict since it arises from sudden biomechanical events rather than a gradual decline in ARS score.

The ARS score on April 11 was approximately 60, placing the athlete within the standard condition range. Following the injury, the score rose steadily and eventually reached 95, which falls within the good condition threshold and reflects optimal readiness. This upward trend was driven by reduced physical strain, improved wellness from sufficient rest, and declined injury risk. 

Overall, while not all injuries can be predicted through fatigue and performance metrics, the post-injury period highlights the effectiveness of the {\THESYSTEM} system in capturing recovery dynamics and providing a clear signal of the athlete’s return to stable condition.

\subsection{Case Study 2: Cumulative Match Load}
\begin{figure}[h]
\centering
\includegraphics[width=0.45\textwidth]{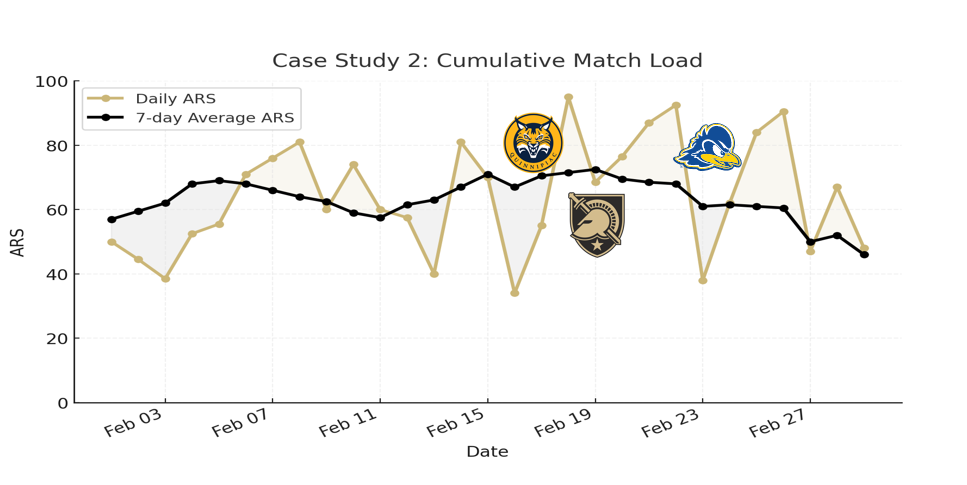}
\caption{ARS tracking during match series (February 2024). The plot shows a gradual decrease in ARS following three consecutive matches (indicated by university logos), reflecting accumulated fatigue.}
\label{fig:case_study_2}
\end{figure}

Figure~\ref{fig:case_study_2} illustrates the ARS response to a period of intense competition in February 2024, featuring three consecutive matches against different institutions.

The three consecutive matches led to a marked decline in ARS, effectively capturing the cumulative fatigue effect of repeated high-intensity competition. The player’s score dropped from about 70 before the series to roughly 40 afterward, shifting from the standard condition range into the high-risk zone. 
This case demonstrates that ARS is not only an indicator of potential injury risk but also a reflection of overall physical condition and athlete wellness, providing valuable guidance for both coaches and players during intensive competitions.

\section{Conclusions}
\label{conclusion}
In this study, a two-stage multimodal prediction framework, {\THESYSTEM}, is designed to monitor athlete wellness and estimate near-term injury risk for tennis athletes through the Athlete Readiness Score (ARS). In the integration stage, the ARS is derived using supervised learning based on four meaningful factors obtained by applying various machine learning techniques to multimodal data during the separating stage. Our defined ARS within {\THESYSTEM} effectively captures the overall condition of tennis athletes, as well as injury-prone body areas. The effectiveness of the framework is demonstrated through two case studies. As a pilot study, data were collected from nine collegiate tennis players over one semester using online questionnaires, wearable devices, professional testing protocols, and video recordings; consequently, the generalizability of the current results should be interpreted within this scope.

We note several limitations of the current work. The playing-style weighting factors are heuristic values informed by domain expertise rather than learned from data and should be treated as preliminary. The limited cohort size (nine players) restricts statistical power and the ability to capture the full diversity of injury patterns. Future validation with larger and more diverse populations is necessary to confirm the robustness of the framework.

Future work will focus on implementing the framework as an intelligent tennis coach software, providing personalized advice using large language models and promoting it to tennis players at all levels.

\section*{Acknowledgment}
This work involved human subjects in its research. Approval of all ethical and experimental procedures and protocols was granted by the Monmouth University Institutional Review Board under Application No. FA2584, with an exemption determination from federal regulations. All participants were informed about the study procedures and provided written informed consent.

\section{About Authors}
\textbf{Weihao Qu} (wqu@monmouth.edu) is an assistant professor of computer science at Monmouth University, West Long Branch, NJ 07764, USA. His research interests include formal verification and artificial intelligence education.

\textbf{Dongyang Wang} (s1382037@monmouth.edu) is pursuing a master's degree in computer science at Monmouth University, West Long Branch, NJ 07764, USA. His research interests include artificial intelligence and data science applications in healthcare and finance.

\textbf{Ling Zheng} (lzheng@monmouth.edu) is an associate professor of computer science at Monmouth University, West Long Branch, NJ 07764, USA. Her research interests include biomedical informatics.

\textbf{Francisco E. Alvarez} (s1365567@monmouth.edu) earned a master's degree in software engineering from Monmouth University in 2024, West Long Branch, NJ 07764, USA.

\textbf{Shobharani Polasa} (s1365603@monmouth.edu) earned a master's degree in data science from Monmouth University in 2024, West Long Branch, NJ 07764, USA.

\textbf{Jiacun Wang} (jwang@monmouth.edu) is a professor of computer science at Monmouth University, West Long Branch, NJ 07764, USA. His research interests include machine learning, formal methods, and discrete event systems. He is the founding editor in chief of the \emph{International Journal of AI and Green Manufacturing} and an associate editor of \emph{IEEE Transactions on Systems, Man, and Cybernetics: Systems}, \emph{IEEE/CAA Journal of Automatica Sinica}, and \emph{IEEE Systems, Man, \& Cybernetics Magazine}.

\bibliographystyle{IEEEtran}
\bibliography{main}

\end{document}